\documentclass[letterpaper, 10 pt, conference]{ieeeconf}  %

\IEEEoverridecommandlockouts                              %

\usepackage[backend=biber,
            hyperref=true,
            url=false,
            isbn=false,
            doi=false,
            backref=false,
            style=ieee,
            natbib=true,%
            mincitenames=1,
            maxcitenames=1,
            citestyle=numeric-comp,
            sorting=nyt,%
            block=none]{biblatex}

\usepackage{amsmath} %
\usepackage{amssymb}  %
\usepackage{graphicx}
\usepackage{xcolor}
\usepackage{algorithm}
\usepackage[noend]{algorithmic}
\usepackage{listings}
\usepackage{xspace}
\usepackage{url}
\usepackage[font={footnotesize}]{caption}
\usepackage[caption=false, font=footnotesize]{subfig}
\usepackage{booktabs} %
\usepackage{tikz}
\usetikzlibrary{positioning,shapes.geometric,arrows,fit,shapes.symbols,svg.path}

\pgfdeclarelayer{background}
\pgfdeclarelayer{foreground}
\pgfsetlayers{background,main,foreground}

\usepackage{autolabtools}

\newcommand{\algname}{FogROS\xspace}

\tikzstyle{rosnode}=[
    rectangle,
    rounded corners,
    text width=42pt,
    minimum height=20pt,
    fill=green!50,
    align=center,
    draw=black]
\tikzstyle{rostopic}=[
    inner sep=0pt,
    trapezium,
    trapezium left angle=75,
    trapezium right angle=105,
    text width=24pt,
    minimum height=20pt,
    fill=blue!30,
    align=center,
    draw=black]
    
\tikzstyle{onrobot}=[
    rounded corners,
    draw]

\tikzstyle{onroscloud}=[
    rounded corners,
    fill=black!5,
    draw]
    
\tikzstyle{oncloud}=[
    cloud,
    cloud puffs=7,
    cloud ignores aspect,
    draw]

\tikzset{pics/cloud/.style={code={
    \draw (-1.5*#1,0.1*#1) circle [radius=1.1*#1];
    \draw ( 1.5*#1,0) circle [radius=#1];
    \draw ( 0,#1) circle [radius=1.25*#1];
    \draw (-1.5*#1,-#1) -- (1.5*#1,-#1);

    \fill [white] (-1.5*#1,0.1*#1) circle [radius=1.1*#1-0.2pt];
    \fill [white] ( 1.5*#1,0) circle [radius=#1-0.2pt];
    \fill [white] ( 0,#1) circle [radius=1.25*#1-0.2pt];
    \fill [white] (-1.5*#1,-#1+0.2pt) rectangle (1.5*#1,0);
}}}

\title{\LARGE \bf
FogROS: An Adaptive Framework \\ for Automating Fog Robotics Deployment
}

\author{%
Kaiyuan (Eric) Chen$^{1}$, 
Yafei Liang$^{1}$,
Nikhil Jha$^{1}$,
Jeffrey Ichnowski$^{1}$,
Michael Danielczuk$^{1}$, \\
Joseph Gonzalez$^{1}$,
John Kubiatowicz$^{1}$,
Ken Goldberg$^{1,2}$%
\thanks{$^{1}$Department of Electrical Engineering and Computer Science,
        University of California, Berkeley, CA, USA 
        $^{2}$Department of Industrial Engineering \& Operations Research,
        University of California, Berkeley, CA, USA {\tt\small \{kych, debbieliang-123, nikhiljha, jeffi, mdanielczuk, jegonzal, kubitron, goldberg\}@berkeley.edu}}%
}

\begin{document}

\maketitle
\thispagestyle{empty}
\pagestyle{empty}

\begin{abstract}
As many robot automation applications increasingly rely on multi-core processing or deep-learning models, cloud computing is becoming an attractive and economically viable resource for systems that do not contain high computing power onboard. Despite its immense computing capacity, it is often underused by the robotics and automation community due to lack of expertise in cloud computing and cloud-based infrastructure. Fog Robotics balances computing and data between cloud edge devices. %
We propose a software framework, \algname, as an extension of the Robot Operating System (ROS), the de-facto standard for creating robot automation applications and components.
It allows researchers to deploy components of their software to the cloud with minimal effort, and correspondingly gain access to additional computing cores, GPUs, FPGAs, and TPUs, as well as predeployed software made available by other researchers.
 \algname %
 allows a researcher to specify which components of their software will be deployed to the cloud and to what type of computing hardware.
We evaluate \algname on 3 examples: (1) simultaneous localization and mapping (ORB-SLAM2),
(2) Dexterity Network (Dex-Net) GPU-based grasp planning,
and (3) multi-core motion planning using a 96-core cloud-based server.
In all three examples, a component is deployed to the cloud and accelerated with a small change in system launch configuration, while incurring additional latency of 1.2\,s, 0.6\,s, and 0.5\,s due to network communication, the computation speed is improved by $2.6\times$, $6.0\times$ and $34.2\times$, respectively.
Code, videos, and supplementary  material can be found  at \url{https://github.com/BerkeleyAutomation/FogROS}.

\end{abstract}

\section{Introduction}

Power, weight, and cost considerations often mean robots do not include computing capabilities capable of running large-scale multi-core CPU-based, graphics processing unit (GPU)-based, field-programmable gate array (FPGA)-based, or tensor processing unit (TPU)-based algorithms.  
For example, a light-weight drone with an attached gripper that uses a GPU-based grasp-planning module to compute grasp points for picking up objects~\cite{backus2014design} or perching~\cite{ramon2019autonomous}, requires access to a GPU that the drone would not have onboard.
While nearby computers can provide the necessary computing capabilities, this practice can be complex to set up, scale, and is prone to over-provisioning.
Instead, we propose a framework based on the \emph{Fog Robotics}~\cite{ichnowski2020fog,tian2017cloud,tanwani2019fog} idea of balancing between the compute available at the edge and in the cloud.
This framework, \emph{FogROS}, is an extension of the Robot Operating System (ROS)~\cite{quigley2009ros} that, with minimal effort, allows researchers to deploy components of their software to the cloud, and correspondingly gain access to additional computing cores, GPUs, FPGAs, and TPUs, as well as predeployed software made available by other researchers. %
\begin{figure}
    \centering
    \subfloat[\algname Application on VPC]{%
      \begin{tikzpicture}[node distance=6pt,>=stealth',font=\footnotesize,
   rosblock/.style={draw, rectangle, rounded corners, minimum height=24pt, text width=64pt, align=center, inner sep=0}]
   
\node [rosblock, text width=0.4in, fill=blue!30] (rosmaster) { ROS \\ Master };
\node [rosblock, fill=green!30, below=6pt of rosmaster.south east, anchor=north east] (sensornode) { Sensor \\ Node };
\node [rosblock, fill=green!30, below=6pt of sensornode.south east, anchor=north east] (controlnode) { Control \\ Node };

\node [anchor=north west, inner sep=0] (robotlabel) at (sensornode.west |- rosmaster.north) { \bf Edge };

\node [rosblock, fill=green!30, right=90pt of sensornode.north east, anchor=west] (graspnode) { Grasp Planner \\ Node };

\node [rosblock, fill=green!30, below=18pt of graspnode] (motionnode) { Motion Planner \\ Node\quad. };
\node [above=6pt of graspnode.north east,anchor=south east,inner sep=0] (vpclabel) { \bf VPC };

\draw [thick,->,color=black] (graspnode) -- node [inner sep=2pt, rounded corners, color=black, fill=white, fill opacity=0.5, text opacity=1, align=center] {Grasp Target Topic} (motionnode);
\path [thick,->,color=black] (sensornode) edge [bend left=10] node [color=black, align=center] (sensortopic) {Sensor \\ Topic} (graspnode);
\draw [thick,->,color=black] (motionnode) -- node [color=black, align=center] (plantopic) {Plan \\ Topic} (controlnode);

\begin{scope}[shift=(sensortopic.east),xshift=2pt, color=black!70]
  \draw [rounded corners=1.5pt,very thick] (-2.5pt,0) rectangle (2.5pt,6pt);
  \fill [rounded corners=1.5pt] (-4pt,-4pt) rectangle (4pt,3pt);
  \fill [color=white] (0,0.5pt) circle[radius=1pt];
  \fill [color=white] (0,0.5pt) -- (-1pt,-2.5pt) -- (1pt,-2.5pt) -- cycle;
\end{scope}

\begin{scope}[shift=(plantopic.east),xshift=2pt, color=black!70]
  \draw [rounded corners=1.5pt,very thick] (-2.5pt,0) rectangle (2.5pt,6pt);
  \fill [rounded corners=1.5pt] (-4pt,-4pt) rectangle (4pt,3pt);
  \fill [color=white] (0,0.5pt) circle[radius=1pt];
  \fill [color=white] (0,0.5pt) -- (-1pt,-2.5pt) -- (1pt,-2.5pt) -- cycle;
\end{scope}

\node [draw, text=white, fill=green!50!black, anchor=south east, inner sep=2pt, shift={(-2pt,2pt)}, rounded corners=2pt, font=\scriptsize] at (graspnode.south east) { \bf GPU };

\foreach \x/\y in {5/3.5,4/3,3/2.5,2/2}
  \node [draw, text=white, fill=blue!70, anchor=south east, inner sep=2pt, shift={(-\x pt,\y pt)}, rounded corners=2pt, font=\scriptsize] at (motionnode.south east) { \bf CPU };

\begin{pgfonlayer}{background}

  \node [fit=(rosmaster)(controlnode),draw=black,rounded corners,fill=black!10] {};
  \node [fit=(vpclabel)(graspnode)(motionnode),draw=black,rounded corners,fill=black!10] (vpc) {};
  
  \path [<->,densely dotted] (rosmaster.210) edge[bend right=50] (sensornode.90);
  \path [<->,densely dotted] (rosmaster.west) edge[bend right=45] (controlnode);
  \path [<->,densely dotted] (rosmaster) edge[bend left=10] (graspnode);
  \path [<->,densely dotted] (rosmaster) edge[bend left=20] (motionnode);

  \path [draw]
       ([shift=(210:8pt)]vpc.210)
       to [bend left=80]
       ([shift=(160:8pt)]vpc.160)
       to [bend left=80]
       ([shift=(140:12pt)]vpc.140)
       to [bend left=60]
       ([shift=(110:6pt)]vpc.110)
       to [bend left=60]
       ([shift=(70:8pt)]vpc.70)
       to [bend left=60]
       ([shift=(50:8pt)]vpc.50)
       to [bend left=60]
       ([shift=(0:8pt)]vpc.0)
       to [bend left=60]
       ([shift=(320:8pt)]vpc.320)
       to [bend left=60]
       ([shift=(290:6pt)]vpc.290)
       to [bend left=60]
       ([shift=(260:8pt)]vpc.260)
       to [bend left=60]
       ([shift=(240:8pt)]vpc.240)
       to [bend left=60]
       cycle;
\end{pgfonlayer}

\end{tikzpicture}%
    } \\[-2pt]
    \subfloat[\algname Application with Proxy]{%
      \begin{tikzpicture}[node distance=6pt,>=stealth',font=\footnotesize,
   rosblock/.style={draw, rectangle, rounded corners, minimum height=24pt, text width=64pt, align=center, inner sep=0}]
   
\node [rosblock, text width=0.4in, fill=blue!30] (rosmaster) { ROS \\ Master };
\node [rosblock, fill=green!30, below=6pt of rosmaster.south east, anchor=north east] (sensornode) { Sensor \\ Node };
\node [rosblock, fill=green!30, below=6pt of sensornode.south east, anchor=north east] (controlnode) { Control \\ Node };

\node [rosblock, text width=0.35in, fill=black!20, right=6pt of sensornode.east, yshift=-16pt] (robotproxy) { Proxy \\ Node };

\node [rosblock, text width=0.35in, fill=black!20, right=26pt of robotproxy] (cloudproxy) { Proxy \\ Node };

\node [anchor=north west, inner sep=0] (robotlabel) at (sensornode.west |- rosmaster.north) { \bf Edge };

\node [rosblock, fill=green!30, right=90pt of sensornode.north east, anchor=west,yshift=-12pt] (graspnode) { Grasp Planner \\ Node };

\node [rosblock, fill=green!30, below=6pt of graspnode] (motionnode) { Motion Planner \\ Node\quad. };
\node [above=6pt of graspnode.north east,anchor=south east,inner sep=0,yshift=12pt] (vpclabel) { \bf Cloud };

\node [rosblock, text width=0.4in, fill=blue!30, anchor=west] (rosmastercloud) at (rosmaster.east -| graspnode.north west) { ROS \\ Master };

\path [thick,color=black] (sensornode) edge [->,bend left=20] node [color=black, align=center, shift={(12pt,4pt)}] (sensortopic) {\scriptsize Sensor \\ Topic} (robotproxy.120);
\draw [thick,color=black] (motionnode) edge [->,bend left=20] node [color=black, align=center, shift={(-12pt,-4pt)}] (plantopic) {\scriptsize Plan \\ Topic} (cloudproxy.300);

\path [thick,color=black] (cloudproxy.60) edge [->,bend left=20] node [color=black, align=center, shift={(-12pt,4pt)}] (plandtopiccloud) {\scriptsize Sensor \\ Topic} (graspnode);
\draw [thick,color=black] (robotproxy.240) edge [->,bend left=20] node [color=black, align=center, shift={(12pt,-4pt)}] (plantopiccloud) {\scriptsize Plan \\ Topic} (controlnode);

\draw [very thick, color=black] (robotproxy) edge [->,bend left=15] (cloudproxy);
\draw [very thick, color=black] (cloudproxy) edge [->,bend left=15] (robotproxy);

\node [draw, text=white, fill=green!50!black, anchor=south east, inner sep=2pt, shift={(-2pt,2pt)}, rounded corners=2pt, font=\scriptsize] at (graspnode.south east) { \bf GPU };

\foreach \x/\y in {5/3.5,4/3,3/2.5,2/2}
  \node [draw, text=white, fill=blue!70, anchor=south east, inner sep=2pt, shift={(-\x pt,\y pt)}, rounded corners=2pt, font=\scriptsize] at (motionnode.south east) { \bf CPU };

\begin{pgfonlayer}{background}

  \node [fit=(rosmaster)(controlnode)(robotproxy)(plantopiccloud),draw=black,rounded corners,fill=black!10] {};
  \node [fit=(vpclabel)(graspnode)(motionnode)(cloudproxy.100),draw=white,rounded corners] (vpc) {};
  
  \path [<->,densely dotted] (rosmaster.210) edge[bend right=50] (sensornode.90);
  \path [<->,densely dotted] (rosmaster.west) edge[bend right=45] (controlnode);
  \path [<->,densely dotted] (rosmaster) edge[out=320,in=140] (robotproxy);
  
  \path [<->,densely dotted] (rosmastercloud) edge[bend left=40] (graspnode);
  \path [<->,densely dotted] (rosmastercloud) edge[bend left=60] (motionnode);
  \path [<->,densely dotted] (rosmastercloud) edge[out=220,in=40] (cloudproxy);

  \path [draw]
       ([shift=(210:8pt)]vpc.210)
       to [bend left=80]
       ([shift=(160:8pt)]vpc.160)
       to [bend left=80]
       ([shift=(140:12pt)]vpc.140)
       to [bend left=60]
       ([shift=(110:6pt)]vpc.110)
       to [bend left=60]
       ([shift=(70:8pt)]vpc.70)
       to [bend left=60]
       ([shift=(50:8pt)]vpc.50)
       to [bend left=60]
       ([shift=(0:8pt)]vpc.0)
       to [bend left=60]
       ([shift=(320:8pt)]vpc.320)
       to [bend left=60]
       ([shift=(290:6pt)]vpc.290)
       to [bend left=60]
       ([shift=(260:8pt)]vpc.260)
       to [bend left=60]
       ([shift=(240:8pt)]vpc.240)
       to [bend left=60]
       cycle;
\end{pgfonlayer}

\end{tikzpicture}%
    }
    \caption{{\bf \algname Applications and Communications}.  ROS applications using \algname run nodes on the cloud or on an edge computer with a small change to the configuration script.  \algname sets up the cloud computers and the secure communication channels between the edge computer and the cloud, either through a virtual private cloud (VPC) or through transparent proxying.%
    }
    \label{fig:proxy_overview}
\end{figure}
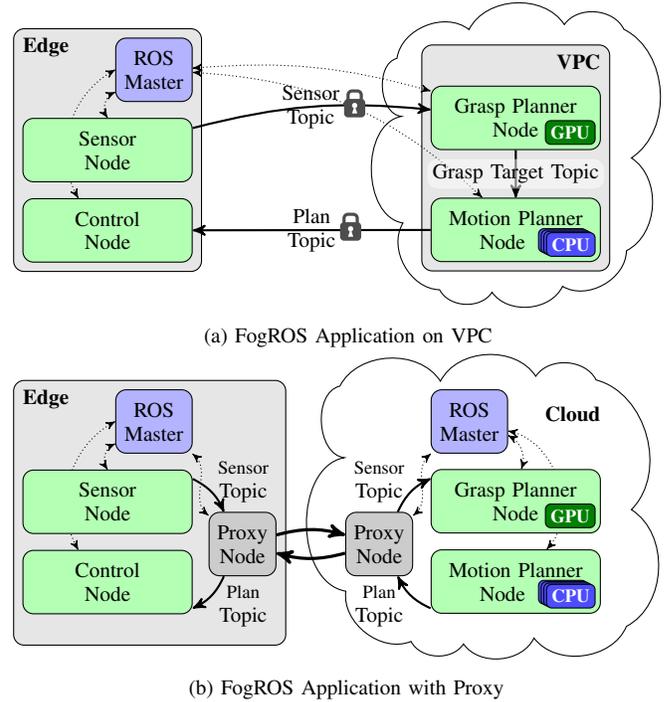

ROS, at its core, is a platform in which software components (\emph{nodes}) communicate with each other via a publication/subscription (pub/sub) system.  Individual nodes can publish messages to named \emph{topics} and subscribe to other named topics to get messages published by other nodes.  In practice, these nodes all run on the robot and perhaps a nearby computer.  For example, on a robot, a sensor node publishes to a sensor topic, a 
planning node subscribes to the sensor topic and compute a plan based on the sensor messages, and then publishes messages that another node uses to execute the plan (Fig.~\ref{fig:proxy_overview}).

With \algname, a researcher can use the same code, and make a small change to a configuration file to select components of the edge computer software to deploy to cloud-based computers.  On launch, \algname  provisions a cloud-based computer, deploys the nodes to it, and then transparently passes the pub/sub communication between the edge computer and the cloud.  The only observable differences are: (1) the pub/sub latency increases, and
(2) the cloud-deployed components can compute faster given the additional computing resources.
The increased latency means that not all components will benefit from being deployed to the cloud, in particular, any component with real-time requirements (e.g., a motor controller) or any component that requires little computing power, should not be deployed to the cloud.
On the other hand, for many applications, the increased computation speed may enable new robot capabilities, speed up tasks, and allow for higher accuracy in tasks such as object detection or segmentation due to the use of larger models.%

\algname also supports launching pre-built automation container images.  These container images contain all the software and dependencies required to run a program. To date, many academic and industrial open-source communities leverage container services, such as Docker~\cite{docker}, to distribute their applications. \algname %
can
deploy robot automation containers to the cloud without explicitly configuring the environment and hardware, facilitating ease of containerized software reuse.

This paper makes three contributions:
    (1) \algname, an open-source extension to ROS that allows user-friendly and adaptive deployment of software components to cloud-based computers;
    (2) a method to pre-deploy containerized \algname software that allows commonly-used software to be quickly integrated into applications;
    and
    (3) application examples evaluating the performance of \algname deployment.

\subsection{Design Principles}

\algname aims to adhere to the following design principles:

\paragraph{Transparent to software} \algname should preserve ROS abstractions and interfaces. Applications should notice no difference between cloud-deployed and on-board nodes (other than the latency of message processing). 

\paragraph{Flexible computing resources} Different nodes require different computing capabilities. Some nodes benefit from additional computing cores, while others benefit from access to a GPU. \algname should make selecting the appropriate configuration simple.

\paragraph{Minimal configuration required} Running software nodes on a cloud-based computer should be as easy as running them on the edge computer.

\paragraph{Pre-deployed nodes} Some useful nodes require extensive setup, installation of a dependency structure, and may have conflicting dependency versions.  \algname should make it possible to %
use pre-deployed containerized software through configuration. %

\paragraph{Flexible Networking} Different networking options may have different availability, performance, setup time, and costs associated.  \algname should allow the user to select the networking options best suited to their application.

\paragraph{Security and Isolation} The communication between cloud and on-board nodes should be secure, and \algname should close ports that expose software to compromise.
\section{Related Work}

Cloud computing has emerged as an attractive and economically viable~\cite{ichnowski2020economic} resource to offload computation for robot automation systems with minimal onboard computing power. \citet{kehoe2015survey} survey the capabilities, research potential, and challenges of cloud robotics, as well as applications such as grasp planning, motion planning, and collective robot learning, that might benefit from the computational power of the cloud. Grasp planning and motion planning have both shown to be amenable to cloud computation. \citet{kehoe2013cloud}, \citet{tian2017cloud}, and \citet{li2018dex} generate robot grasp poses in the cloud by implementing parallelizable Monte-Carlo sampling of grasp perturbations~\cite{kehoe2012estimating,kehoe2012toward,kehoe2014cloud} while \citet{mahler2016privacy} explore cloud grasp pose computation that maintains privacy of proprietary geometries. In motion planning, \citet{lam2014path} introduce path planning as a service (PPaaS) for on-demand path planning in the cloud and use Rapyuta to share plans among robots. \citet{bekris2015cloud} and \citet{ichnowski2016cloud} both devise methods for splitting motion planning computation between the cloud and the edge computer~\cite{ichnowski2020fog,anand2021serverless}. In addition to providing computing resources, the cloud can also facilitate sharing and benchmarking of algorithms and models between edge computers~\cite{tanwani2020rilaas} for grasping, motion planning, or computer vision.

To leverage cloud resources, many in academia and industry endeavor to connect edge computers to the cloud. Example approaches include using SSH port forwarding~\cite{hajjaj2017establishing} or VPN-based proxying~\cite{lim2019cloud} to support unmodified ROS applications to share a single ROS master. \algname builds on these approaches, and adds automation of ROS node deployment to the cloud and a virtual private cloud (VPC), saving time over prior approaches that require manual configuration of network access rules and IP addresses. For example, setting up a VPN-based proxying requires more than 12 steps for configuration and 37 steps for verification~\cite{hajjaj2017establishing}. The complex manual configurations scale poorly and are error-prone. ROSRemote~\cite{pereira2019rosremote} and MSA~\cite{xu2020cloud} replace the ROS communication stack with custom Pub/Sub designs. Although edge computers can communicate with nodes owned by other ROS masters, these systems require heavy code changes to ROS applications. \algname, as an option, leverages rosbridge~\cite{crick2012rosbridge}, an open-source webserver that enables an edge computer to interact with another ROS environment with JSON queries. Given diverse attempts to connect edge computers to the cloud, \citet{wan2016cloud} and \citet{saha2018comprehensive} call for a unified and standardized framework to handle cloud-robot data interactions. \algname aims to be a painless solution to this open issue by allowing unmodified ROS applications to be launched on the cloud with minimal additional configurations.

Sharing a similar vision as \algname, RoboEarth~\cite{waibel2011roboearth} is a successful example where edge computers share information on the cloud. However, in their use cases, edge computers mainly use the shared database on the cloud, and do not benefit from the powerful cloud computing resources. Rapyuta~\cite{mohanarajah2014rapyuta} and AWS Greengrass~\cite{greengrass} provide pipelines to deploy pre-built ROS nodes to edge computers or robots. Both platforms build ROS nodes or Docker images on the cloud, and push the built images to robots that are registered with their platforms. 
\algname considers the reversed direction of Rapyuta and AWS Greengrass. Instead of pushing the computation from cloud to robots, 
\algname is a lightweight platform that allows developers to rapidly prototype applications and gain quick access to extensive computing resources, without conforming to an additional framework. %
\section{Background}

In this section, we provide a brief background on the building blocks of \algname, including (A) cloud-based computing, (B) ROS and its pub/sub system, and (C) how ROS-based robotic systems are configured and launched.

\subsection{Cloud Computing}

Cloud-based computing services, such as Amazon Web Services (AWS), Google Cloud, and Microsoft Azure, offer network accessible computers of various specifications to be rented on a per-time-unit basis.  Setting up a service typically requires a one-time registration and a credit card.  Registered users can setup, reconfigure, turn on, turn off, and tear down virtual computers in the cloud.  This can be done either through a web-browser interface, or programmatically through a network-based application programming interface (API).  Computer configuration options include: amount of memory, amount of processing cores, type and amount of GPUs, and inclusion of custom processing hardware such as field-programmable gate arrays (FPGAs) and tensor processing units (TPUs).
\algname uses the AWS cloud service API to setup a cloud-based computer, deploy ROS and the code, secure network communications, and then run the node.

\subsection{ROS and Pub/Sub}

In ROS, \emph{nodes} (software components) communicate with each other using a pub/sub (publication and subscription) system.  Nodes register as publishers and/or subscribers to named communication channels called \emph{topics}.  Each topic has message type that determines what data is sent over the channel.  For example, a  ROS node that monitors the joint state (e.g., angles) through sensors, would publish messages of type {\tt JointState} on an appropriately named topic, and that topic would only contain {\tt JointState} messages.  When a node publishes a sequence of messages to a topic, all registered subscribers will receive the message in the same sequence they were published.  

Coordination of the publishers and subscribers to topics is maintained by the ROS \emph{Master}~\cite{ROSMaster}.  The ROS Master exposes 
network API
that allows nodes to connect over a network and register/unregister themselves as publishers and subscribers to topics. During the registration process, publishers get the current list of subscribers, and subscribers get the current list of publishers.  Publishers may then connect directly to already-registered subscribers, and subscribers may connect directly to already-registered publishers.

Once publishing and subscribing nodes are directly connected to each other\footnote{As an implementation optimization, ROS nodes on the same machine can communicate via a shared-memory queue, instead of using a network.}, publishing nodes %
serialize message-specific data structure to a sequence of bytes and sends the bytes over the connection.  When subscribing nodes receive the sequence of bytes, they deserialize the bytes to the message-specific data structure and process the message.

However, existing ROS pub/sub communication has several limitations: (1) all the nodes have to share the same master to communicate (inter-master communication is not supported by ROS pub/sub protocol stack, and one has to use out-of-band protocols for communicating across masters); (2) although it is possible to join nodes from multiple machines to share a single master, the communication for ROS is not 
secured,
and users must configure security protocols.%

\subsection{ROS Launch Scripts}

Robot systems can be comprised of a complex graph of nodes communicating with each other via pub/sub.  To consolidate an automation system deployment into a single file, ROS supports a launch configuration file.  This file specifies which nodes are to be launched by code entry point, and allows for optional remapping of topic names (e.g., so that code written to process a standard message type can produce it from a topic with a name not known/specified when the code was written).  Listing  \ref{listing:local} is an example launch script that launches a client node and a server node from the {\tt mpt\textunderscore ros} package locally. 

\begin{lstlisting}[language=XML,caption={ROS Launch Script Example},label=listing:local]
<launch>
    <!-- Run client node -->
    <node name="client" pkg="mpt_ros" type="client" output="screen" />

    <!-- Run server node-->
    <node name="server" pkg="mpt_ros" type="server" output="screen" />
</launch>
\end{lstlisting} 

\algname extends the launch script capabilities to allow the specification of which nodes to deploy in the cloud and on what machine type.

\section{Approach}

To meet the design principles, \algname
(1) extends ROS launch scripts to include an option of where to deploy and run a ROS node, the only place that requires user configurations;
(2) provisions cloud-based computers, securely pushes the code or containers to them, and runs the code;
(3) sets up one of two networking options (VPC or Proxy) to transparently and automatically proxy the pub/sub communication between the edge computer and the cloud;
(4) provides introspection infrastructure for monitoring network conditions;
and
(5) supports launching containerized \algname nodes from pre-built Docker images.

\subsection{Launch Script Extensions} \label{sec:launch_script_extensions}

\algname uses standard ROS launch scripts as the user interface. Users specify which nodes are to be deployed and what type of cloud computing instance is used in the same launch file as the nodes that users want to deploy locally. They can push multiple nodes to the cloud at the same time by providing the path to a separate launch script. \algname parses the launch script, finds and collects all the packages in the script, and pushes them to the cloud computer. As part of the configuration process, users can optionally specify a bash script that installs dependencies outside of the \algname launch process (e.g., mirroring the steps to install dependencies on the edge computer).

Listing~\ref{listing:launch} provides an example of a \algname launch script that serves the same functionality as Listing~\ref{listing:local}, but with the server node running on a cloud computer. Local ROS nodes, such as {\tt client}, are launched as before. With \algname, the user specifies the launch file that contains the server node ({\tt server.launch}), the type of cloud computer ({\tt c5.24xlarge}), and optionally, a setup script ({\tt init.bash}).

\begin{lstlisting}[language=XML,caption={\algname Launch Script Example}, label=listing:launch]
<launch>
    <!-- Run client node as before -->
    <node name="client" pkg="mpt_ros" type="client" output="screen" />

    <!-- Run server node w/ FogROS-->
    <node name="server" pkg="fogros" type="fogros_launch.py" >
        <rosparam>
            instance_type: c5.24xlarge
            launch_file: server.launch
            env_script: init.bash
        </rosparam>
    </node>
</launch>
\end{lstlisting} 

\subsection{Cloud-Computer ROS Nodes}

\begin{figure}
    \centering
    \includegraphics[width=\linewidth]{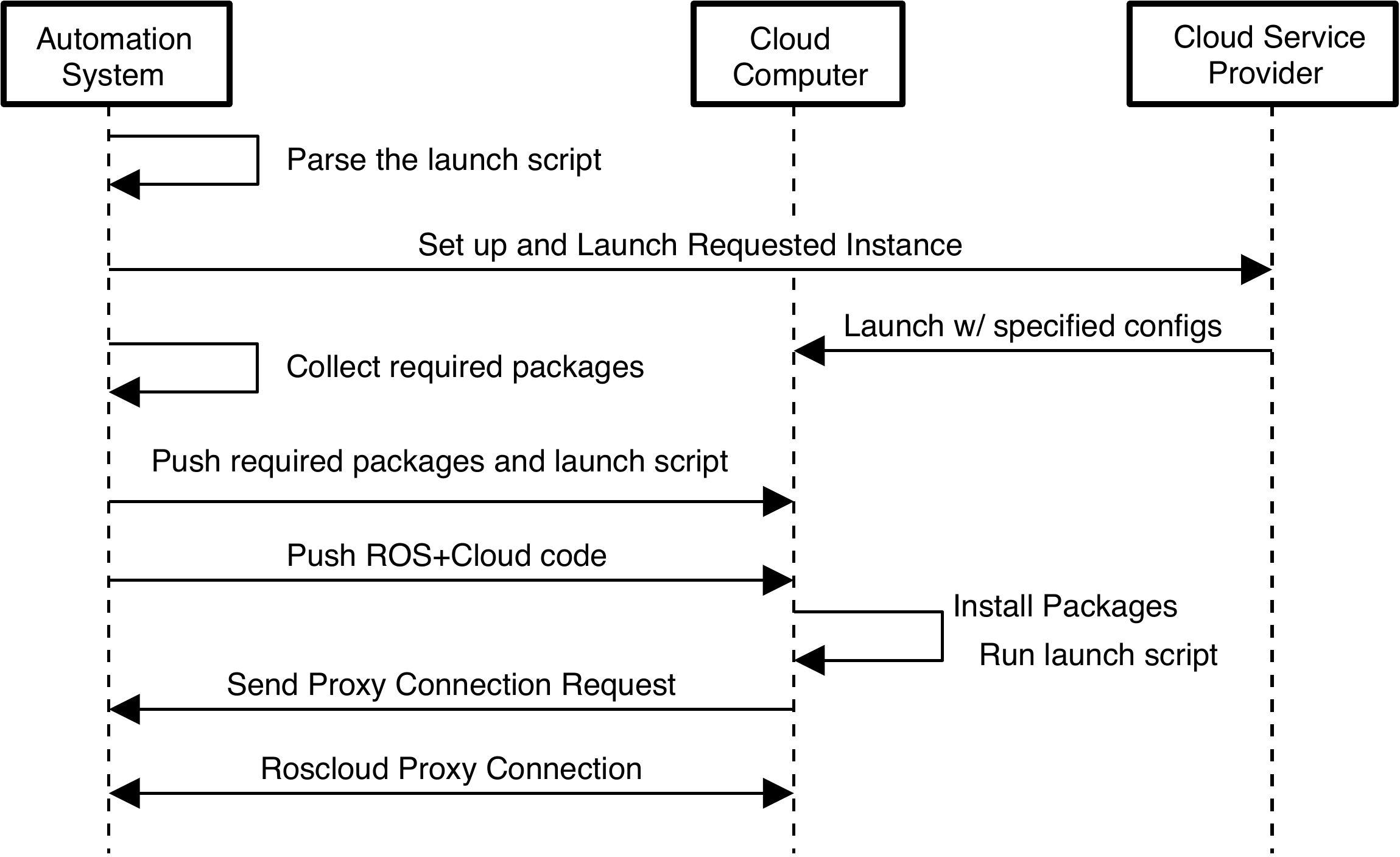}
    \caption{Sequence Diagram of \algname Deployment Process. Users only need to input the launch file, and \algname automates the provisioning, deployment, code execution, and network setup sequence.}
    \label{fig:deployment}
\end{figure}
When \algname launches cloud-based nodes, it performs the following sequence of steps that result in ROS nodes running on a cloud computer with messages being transparently proxied between the edge computer and the cloud computer:
\begin{enumerate}
    \item Provision and start a cloud computer with the capabilities from the launch file and pre-loaded with ROS %
    \item Push code for ROS nodes to the cloud computer %
    \item Run the environment setup script
    \item Set up secure networking (via Proxy or VPC) %
    \item Launch the pushed code
\end{enumerate}

Before \algname provisions a cloud computer, it uses the cloud service provider API to create security rules to set up a secure computing infrastructure suitable for ROS application configuration.
It closes network ports not needed for communication between nodes.
Then it provisions the cloud computer with a specified location and type. To speed up the launching process, \algname specifies an image pre-loaded with the core ROS libraries to run on the cloud computer.  As part of the launch process, \algname generates and installs secure credentials on the cloud computer, and gets its public internet protocol (IP) address. %

Once the computer is started, using the IP address and secure credentials, \algname recursively copies the ROS code to the cloud computer securely over a secure shell (SSH)~\cite{rfc4251} connection, optionally runs the user-specified setup script, and builds the code on the cloud. 
With the code ready to run, \algname then starts the configured secure networking components for VPC (Sec.~\ref{sec:vpc}) or proxying (Sec.~\ref{sec:proxy}), %
and runs the ROS nodes in the cloud.

\subsection{Networking: Virtual Private Cloud}
\label{sec:vpc}

To allow the edge computer and cloud-based computers to communicate securely with each other, \algname automates the setup of a Virtual Private Cloud (VPC). A VPC secures point-to-point communication between cloud computers by assigning private IPs that are only accessible for other nodes within the VPC. 
\algname creates a Virtual Private Network (VPN) between the edge computer and the VPC.  A VPN is a secure network communication channel provided by the operating system.
With this setup, from the perspective of a ROS node, all nodes appear as though they are on the same private network.

\algname automates the setup of the VPC and the VPN when it provisions the cloud computers to run the ROS nodes, by using the cloud service providers API to:
(1) create a VPC instance and a security group for it,
(2) establish credentials for the cloud-computers that will participate in the VPC,
(3) configure the cloud computers to use the VPC for cloud-to-cloud communication,
and
(4) set up a VPN endpoint to which the edge computer will connect.
Once set up, the cloud service provider manages %
the VPC, while FogROS manages the %
VPN.
As part of the setup process, \algname sets a unique private IP address for each of the computers participating, so that the ROS nodes can establish connections between computers.

\subsection{Networking: Pub/Sub Proxying}
\label{sec:proxy}

In addition to VPC networking, \algname also supports a proxied-network option that enable communication between the edge computer and the cloud. This option is available for cases where the VPC solution may be unavailable due to service provider restrictions or costs, or when an additional level of isolation between the edge computer and the cloud is desired.  There are also performance differences (see Section~\ref{sec:evaluation}) when considering the network options, and a user of \algname may wish to measure performance in their application before choosing a suitable option.

In \algname, a proxy consists of two ROS nodes, one running in the edge computer and one running on the cloud. These nodes connect directly to each other via a secure network connection, and register as publishers and subscribers to topics on the ROS Master running on each computer.  When a proxy node receives a message from a subscription, it sends it to the other proxy node, which then publishes it to the subscribers registered on its ROS Master.

\begin{figure}
    \centering
    \includegraphics[width=0.75\linewidth]{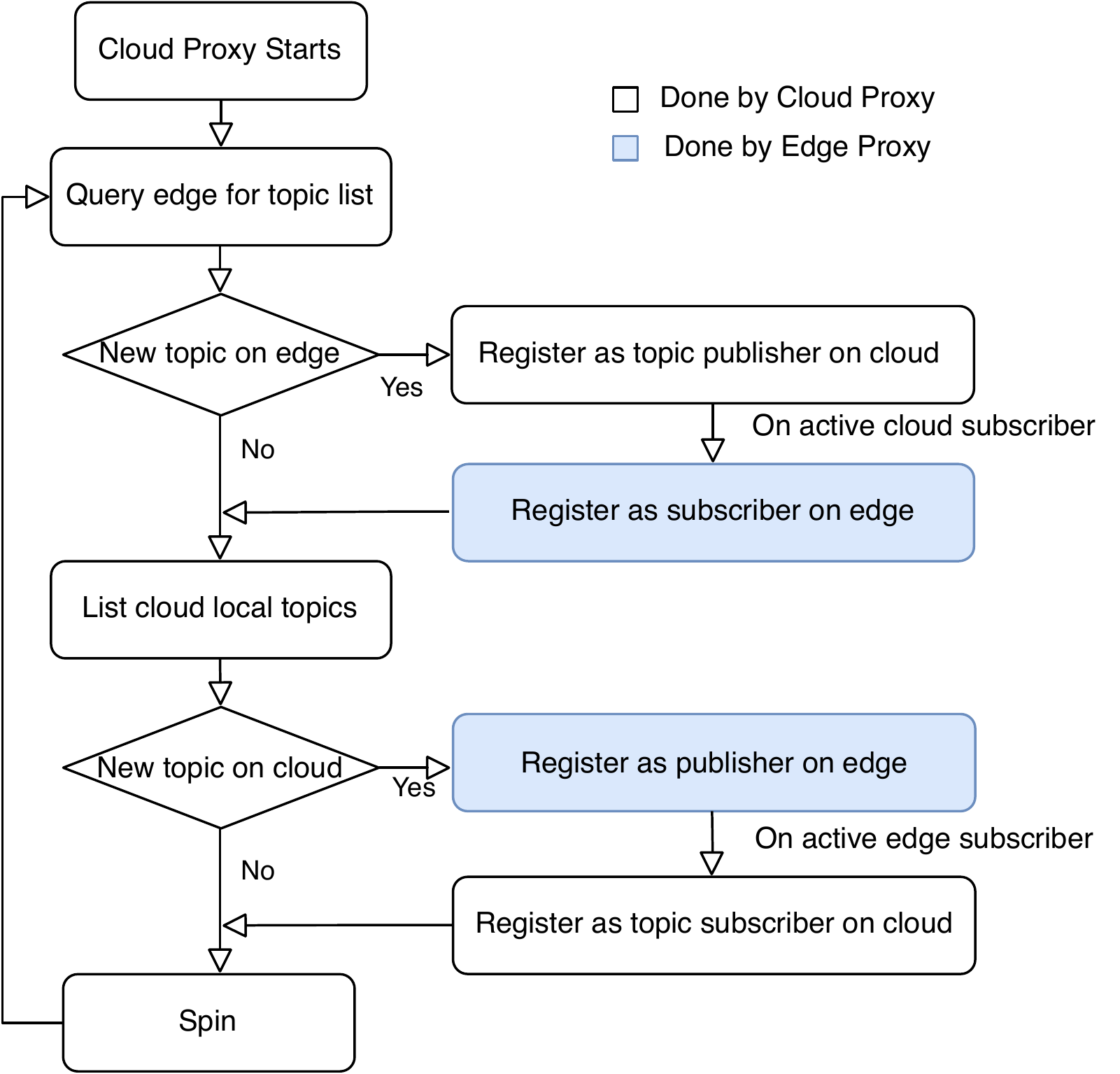}
    \caption{Automatic Pub/Sub Proxy. The proxy tunnels the traffic only if there is an active subscriber.}
    \label{fig:proxy}
\end{figure}

There are two options for \algname to identify topics to proxy: (1) user-specified in the configuration file, or (2) automated.  If topics are specified by the user in the configuration file, \algname subscribes and publishes to the topics specified.  If the user does not specify topics, \algname communicates with the ROS Master on each end and identifies which topics have registered subscribers and publishers.  When a topic has a publisher on one end, and a subscriber on the other, the ROS proxy nodes coordinate with each other to proxy the associated topic (see Fig.~\ref{fig:proxy}). While the automated process is simpler to setup for the developer, it may result in increased setup time as the proxy nodes coordinate the setup of proxied topics, or wasted bandwidth on topics that do not need proxying.

\subsection{Network Monitoring}

With the proxying network option, \algname also provides interfaces to monitor network conditions via ROS topics {\tt /fogros\allowbreak /latency} and {\tt /fogros\allowbreak /throughput} on both the edge computer and the cloud. These interfaces do not introduce additional overhead unless an active subscriber subscribes to them. Users can also inspect and interact with ROS topics with standard ROS tools such as {\tt rostopic}. In addition, \algname provides the same fault tolerance as ROS running locally, where ROS nodes can re-join the pub/sub communication after network interruption. 

\subsection{Pre-Built ROS Nodes}

\algname also supports launching containerized ROS nodes with a similar interface to the \algname launch script extension described in Section~\ref{sec:launch_script_extensions}.
While an increasing number of ROS developers are using pre-built docker images to host ROS nodes, 
this functionality is not natively supported by ROS. %
With \algname, users can specify the name of a publicly-available image on DockerHub as well as the destination machine on which they want to launch it.
\algname then uses a template environment setup script to pull and run the image on the specified machine. It analyzes the machine type and configures the docker run command to match the hardware (e.g., GPU) available on the computer.

Listing \ref{listing:docker} shows an example launch script for a Dex-Net grasp planning node in a docker image. \algname provisions and starts a cloud computer ({\tt g4dn.xlarge}) with a GPU, pulls the {\tt dexnet:gpu} image from DockerHub~\cite{docker}, and %
attaches the docker container to the GPU, and runs it. %

\begin{lstlisting}[language=XML,caption={\algname Docker Container Example}, label=listing:docker]
<launch>
    <!-- Dex-Net Docker Image w/ FogROS -->
    <node name="dexnet" pkg="fogros" type="fogros_docker.py" output="screen">
        <rosparam>
            docker_image: keplerc/dexnet:gpu
            machine_type: g4dn.xlarge
        </rosparam>
    </node>
</launch>
\end{lstlisting}

\section{Evaluation}
\label{sec:evaluation}

\begin{figure*}[t]
    \centering
        \subfloat[VSLAM\label{fig:vslam_nodes}]{%
      \includegraphics[trim=0 50 1227 75, clip, height=120pt]{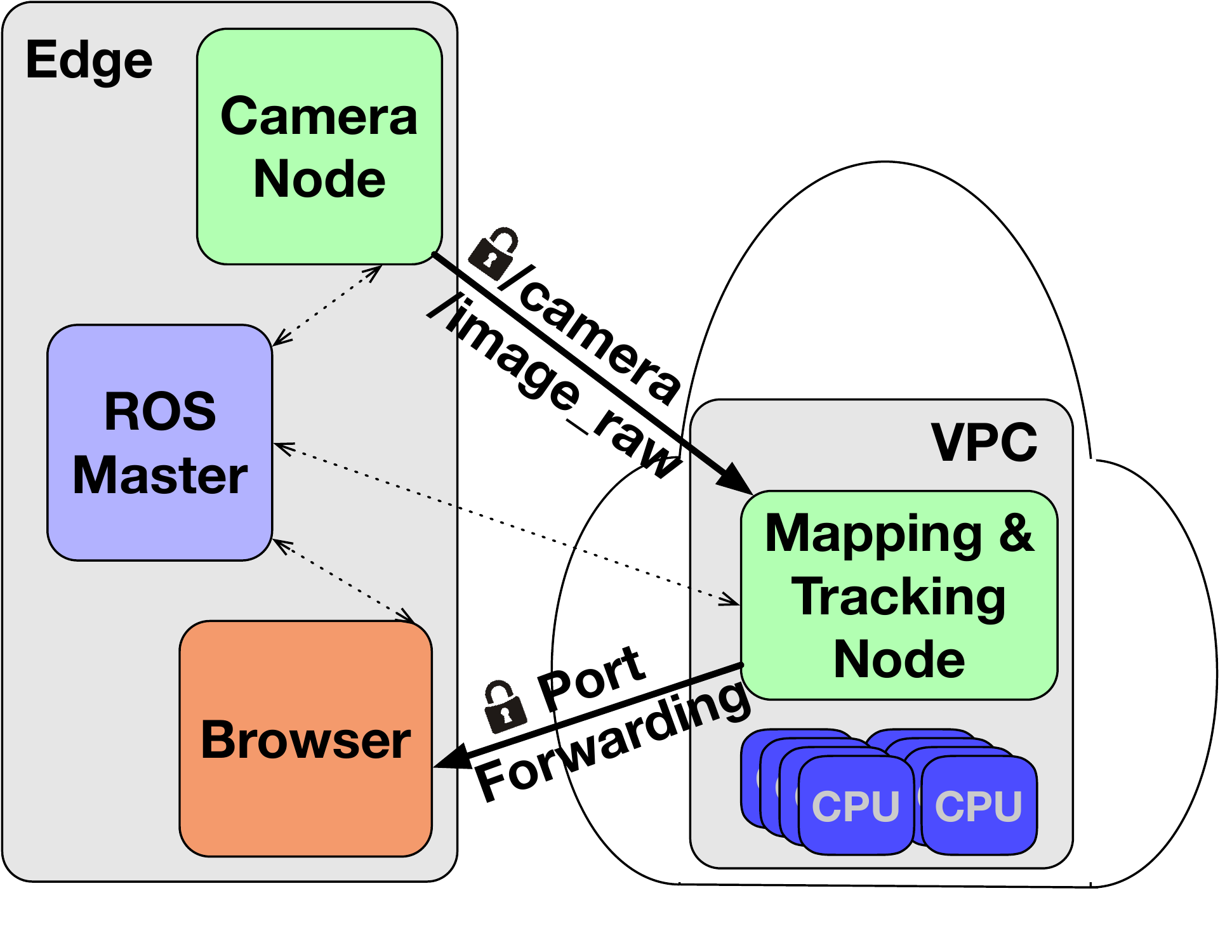}}
      \hfill
    \subfloat[Grasp Planning\label{fig:dn_nodes}]{%
      \includegraphics[trim=554 50 688 75, clip, height=120pt]{image/three_app.pdf}}
      \hfill
    \subfloat[Motion Planning\label{fig:mp_nodes}]{%
      \includegraphics[trim=1100 50 0 75, clip, height=120pt]{image/three_app.pdf}}

    \caption{{\bf Example \algname applications in experiments.}
    In experiments, we run 3 sample applications, each with one node accelerated by cloud computing, with the nodes and topics shown here.  For brevity, we only depict the VPC-based solution here, but we also experiment with the proxy-based solution.
    }
    \label{fig:applications}
\end{figure*}

Here we present three example applications on \algname: (A) visual SLAM, (B) Dex-Net grasp planning and (C) multi-core motion planning.  The nodes, topics, and split between a single-core edge computer with 2GB RAM and the cloud are shown in Fig.~\ref{fig:applications}. In addition to showing the network latency and performance with \algname, we highlight the simplicity and minimal configuration of deploying these applications. %

\subsection{Visual SLAM Service}

\begin{table}[t]
    \centering
    \footnotesize
    \begin{tabular}{@{}lr@{\quad}rr@{\quad}rr@{\quad}r@{}}\toprule
                  & \multicolumn{2}{c}{ Edge}
                  & \multicolumn{4}{c}{\; Cloud \hfill --- \hfill \algname \hfill --- \hfill Network \;} 
                  \\
                  \cmidrule(lr){2-3} \cmidrule(lr){4-5} \cmidrule(l){6-7}
         Scenario & FPS & Create (s)  & FPS & Create (s) &  VPC (s) & Proxy (s) \\
         \midrule
         fr1/xyz & 16.6 & 2.9  & \textbf{25.2} & \textbf{1.2} & 0.4 & 0.6 \\
         fr2/xyz & 14.7 & 2.1  & \textbf{30.2} & \textbf{0.8 }& 0.4 & 0.6 \\
         fr1/desk & 15.6 & 2.8  & \textbf{23.8} & \textbf{2.0} & 0.4 & 0.6 \\
         \bottomrule
    \end{tabular}
    \caption{ SLAM with \algname.  \textup{We benchmark \algname on 3 different visual SLAM scenarios, and record the frame-per-second (FPS) and the latency in creating a new map (Create) in seconds on the local edge computer (using one-core CPU) and a cloud-computer with 36-core CPU. We also record the average network time in seconds for transmitting raw video frames to the cloud computer. \algname demonstrates up to 2.0x improvement on FPS and 2.6x improvement on new map creation time than using only edge computer.}}
    \label{tab:slam_times}
\end{table}

ORB-SLAM2~\cite{mur2017orb} is a visual simultaneous localization and mapping system that uses monocular video input.  In this experiment, a Camera Node publishes a $640\times480$ resolution video with each frame 48\,KiB on average to the cloud (Fig.~\ref{fig:vslam_nodes}). On the cloud an ORB-SLAM2 node subscribes to the video feed~\cite{sturm12iros} and computes a pointcloud map along with the current estimated location within the map, which are sent back to the robot. For more details on the ORB-SLAM2 algorithm, we refer readers to the paper and open-source code available from~\citet{mur2017orb}.

To configure \algname to work with ORB-SLAM2, we build ROS docker images and push them to Dockerhub~\cite{docker}. We wrote a bash script to pull and run the docker image and include its path as in Listing~\ref{listing:launch}; \algname then runs the script when configuring the environment. After initialization, \algname sets up and secures communication between the robot and the cloud SLAM server. 

To evaluate the performance of \algname when deploying ORB-SLAM2, we compare the cloud-deployed performance to an edge-computer-only implementation. We select a 36-core cloud-computer (AWS c4.8xlarge) for the ORB-SLAM2 node, and compare it with ORB-SLAM2 running on a one-core edge computer
We report frames-per-second (FPS) and latency that creates the first map (in seconds)~\cite{nardi2015introducing}. Table~\ref{tab:slam_times} suggests that  cloud-based SLAM 
can achieve higher FPS, meaning that it can aggregate more data and produce higher quality maps in a real-time setting. Cloud-based SLAM also has less latency in generating a new map. 

\subsection{Dex-Net Grasping Service}
Grasp analysis computes the contact point(s) for a robot gripper that maximize grasp reliability---the likelihood of successfully lifting the object given those contact points. To plan grasps on rigid objects in industrial bins using an overhead depth camera, we use an open-source implementation of the fully-convolutional grasp-quality convolutional neural network (FC-GQ-CNN)~\cite{satish2019policy,mahler2019learning} from Dex-Net\cite{mahler2017dex}.  We wrap FC-GQ-CNN in a ROS node and deploy it to the cloud along with pretrained neural-network weights as a Docker image. We refer the reader to \citet{satish2019policy} and \citet{mahler2019learning} for details and code for the neural network and grasping environment.

This node subscribes to 3 input topics containing a scene depth image and mask for objects to be grasped, and a message of type {\tt sensor\_msgs\allowbreak /CameraInfo} containing camera intrinsics. Internally, the node feeds this to FC-GQ-CNN, which outputs a grasp pose and associated estimate of grasp quality. %
The node wraps these outputs, along with the gripper type and coordinates in image space, into a {\tt gqcnn\_ros\allowbreak /GQCNNGrasp} message, and publishes it.%

\begin{figure}
    \centering
    \includegraphics[width=\linewidth]{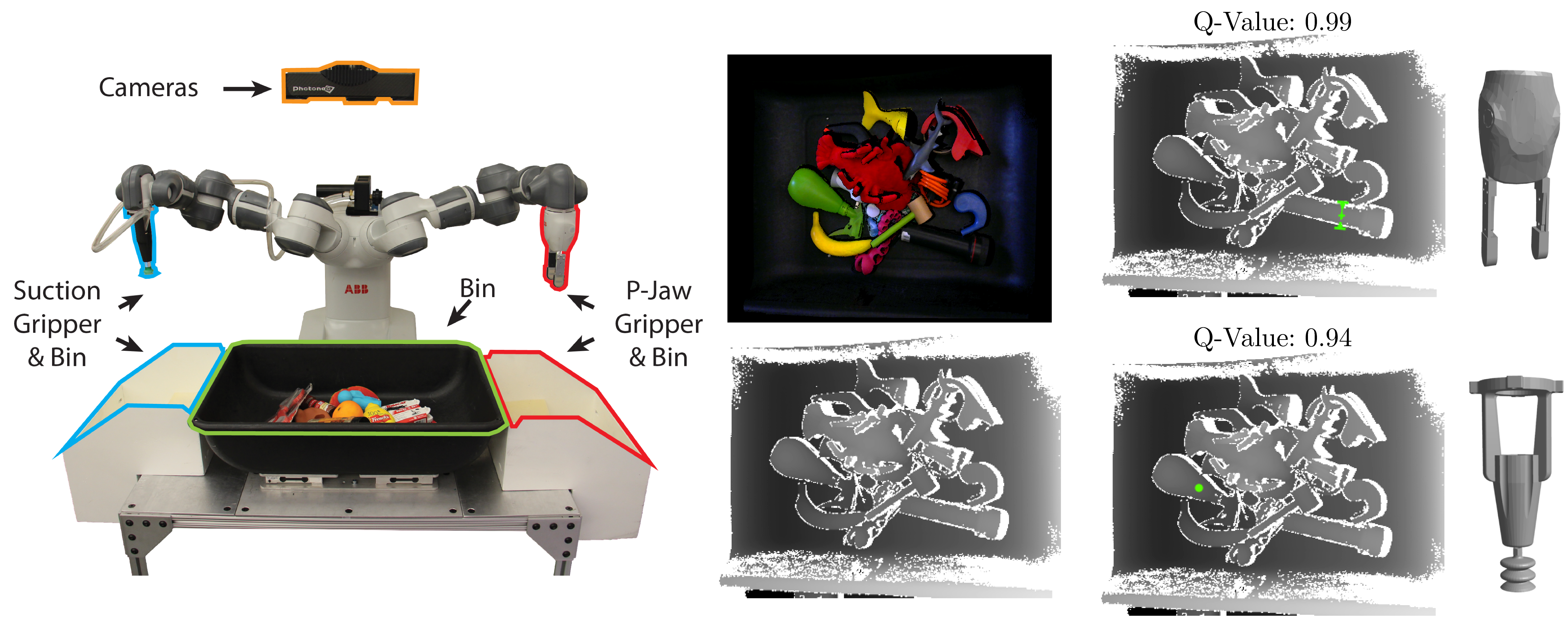}
    \caption{\textbf{Grasp Planning:} A robot with an overhead depth camera and either a suction or parallel jaw gripper (left) must plan a grasp on one of the objects in the bin beneath it given an RGBD image observation (middle). Examples of planned grasps (green) and their q-values are shown for both parallel jaw and suction grippers (right).}
    \label{fig:dex_net_setting}
\end{figure}

While the node can be run both locally or in the cloud, using cloud GPU instances as opposed to a CPU for neural-network inference can greatly reduce computation time. In either case, the node is wrapped inside of a Docker container, reducing the need for resolving dependency issues between deep-learning libraries, CUDA, OS, and ROS versions. 
The pretrained models in the image is intended for a setup similar to that shown in Figure~\ref{fig:dex_net_setting}; variations in camera pose, camera intrinsics, or gripper type may require retraining the underlying model for accurate predictions.

We run \algname with the Dex-Net docker image using the launch file in Listing \ref{listing:docker}. We compare grasp planning times across 10 trials using both the CPU onboard the edge computer and \algname with the Docker images on the cloud. We also show compute times when using a compressed depth image format to transfer images instead of transferring raw images to the cloud directly. For the latter case, images are compressed and decompressed using the \texttt{republish} node from the {\tt image\_transport} ROS package~\cite{ROS_image_transport}. Table~\ref{tab:dn_times} shows the results for both compressed and uncompressed image transport between the nodes.

\begin{table}[t]
    \centering
    \footnotesize
    \begin{tabular}{@{}l@{\quad}c@{\quad}cc@{\quad}cc@{\quad}c@{}}\toprule
                  & Edge & Cloud & \multicolumn{2}{c}{\algname VPC} & \multicolumn{2}{c}{\algname Proxy} \\
                  \cmidrule(lr){4-5} \cmidrule(l){6-7}
         Scenario & Only & Compute & Network & Total & Network & Total \\
         \midrule
         Compressed   & 7.3 &  {0.6} & 0.6 & \textbf{1.2} & 0.8 & \textbf{1.4} \\
         Uncompressed & 7.5 &  {0.6} & 0.7 &  \textbf{1.3} & 0.9 & \textbf{1.5 }\\
         \bottomrule
    \end{tabular}
    
    \caption{Dex-Net Grasp Planning with \algname. \textup{We benchmark Dex-Net on 10 trials, and record the compute time in seconds on the local edge computer (using CPU only), and the total (compute + network) time for using a 4-core cloud computer with a single Nvidia T4 GPU. \algname demonstrates up to 6.0x improvement with VPC and 5.7x improvement with proxy than using only edge server.}}
    \label{tab:dn_times}
\end{table}

\subsection{Multi-Core Motion Planning}

\begin{figure}
    \centering
    \includegraphics[height=46pt]{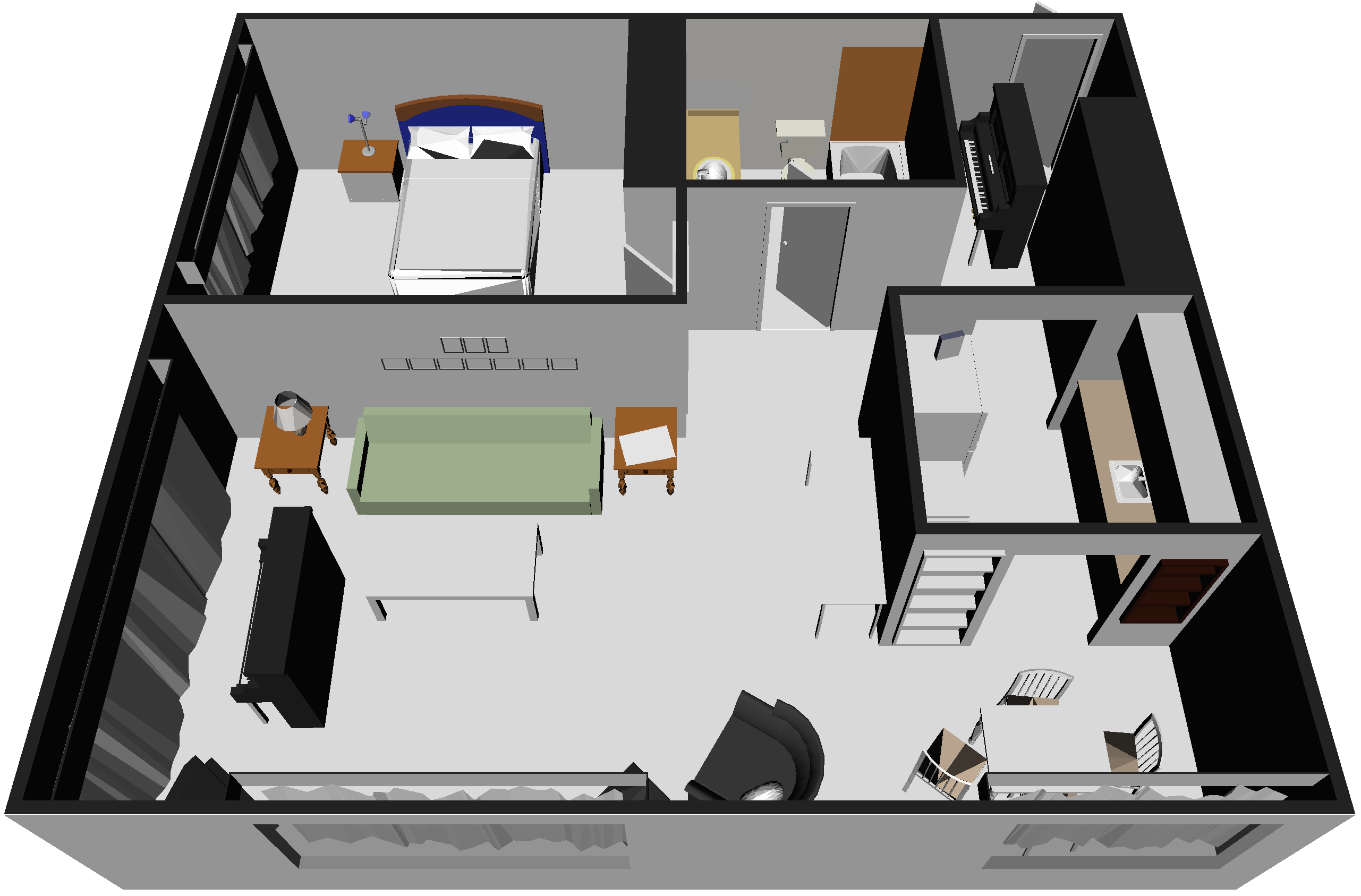}\hfill%
    \includegraphics[height=46pt]{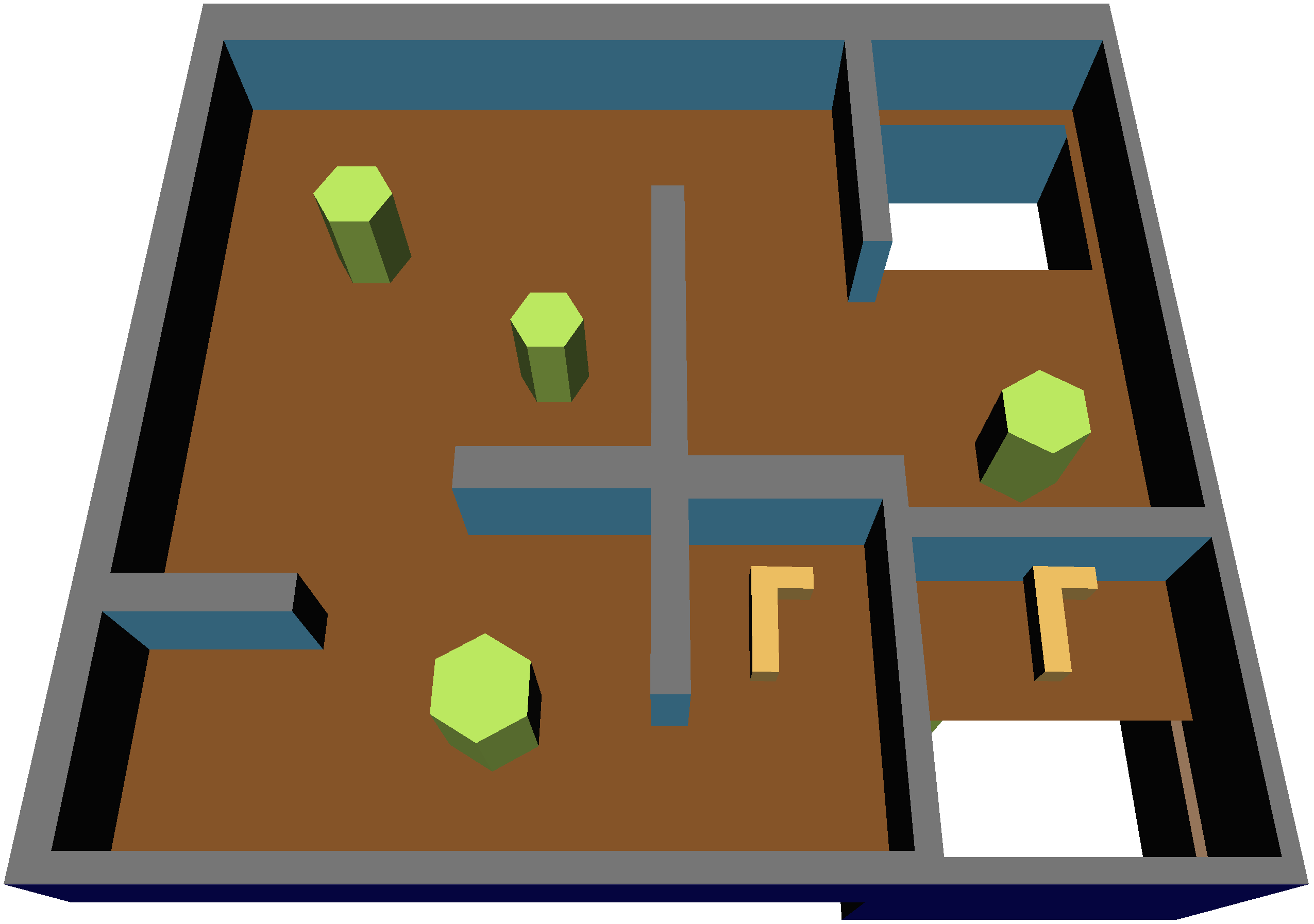}\hfill%
    \includegraphics[height=46pt]{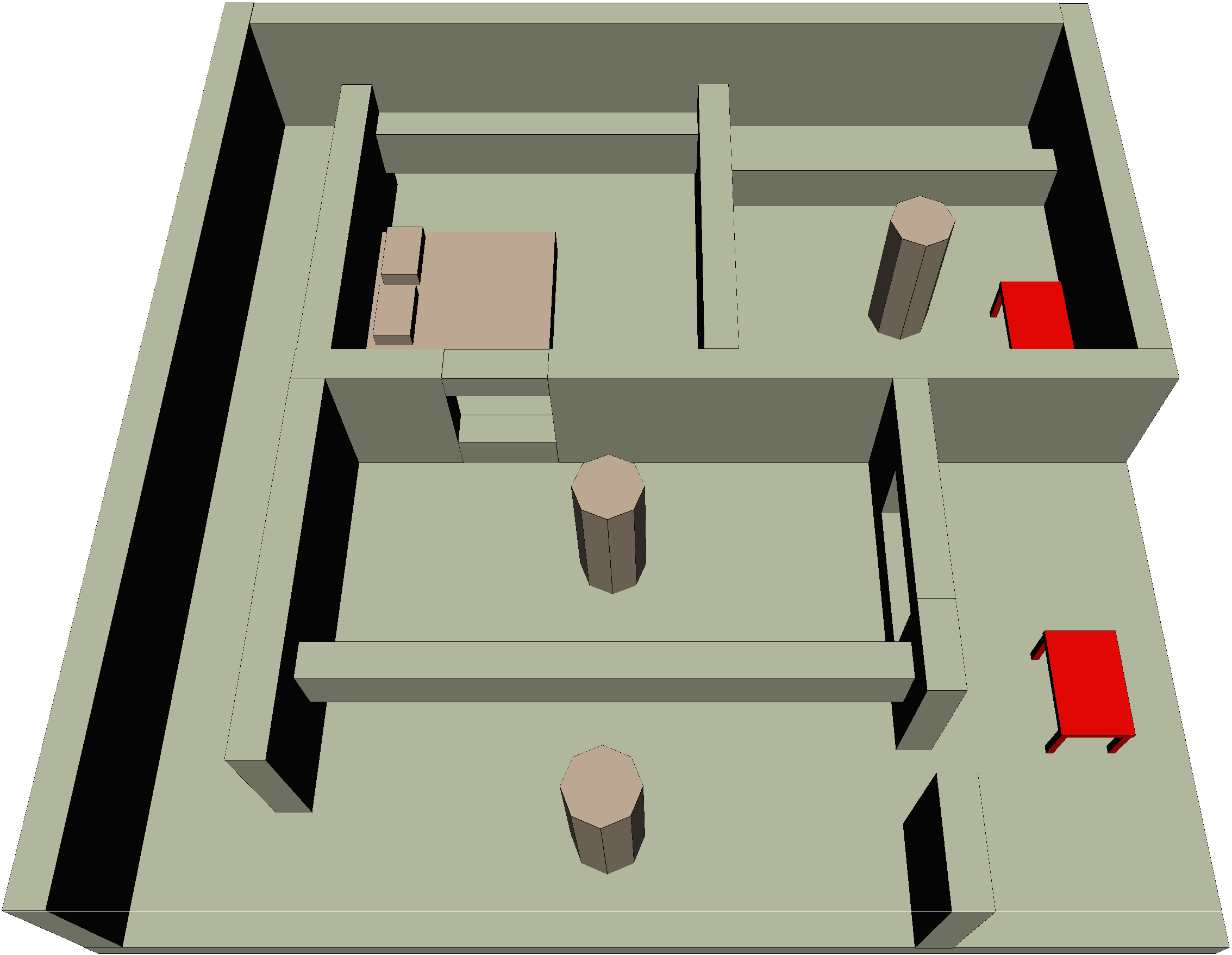}\hfill%
    \includegraphics[height=46pt]{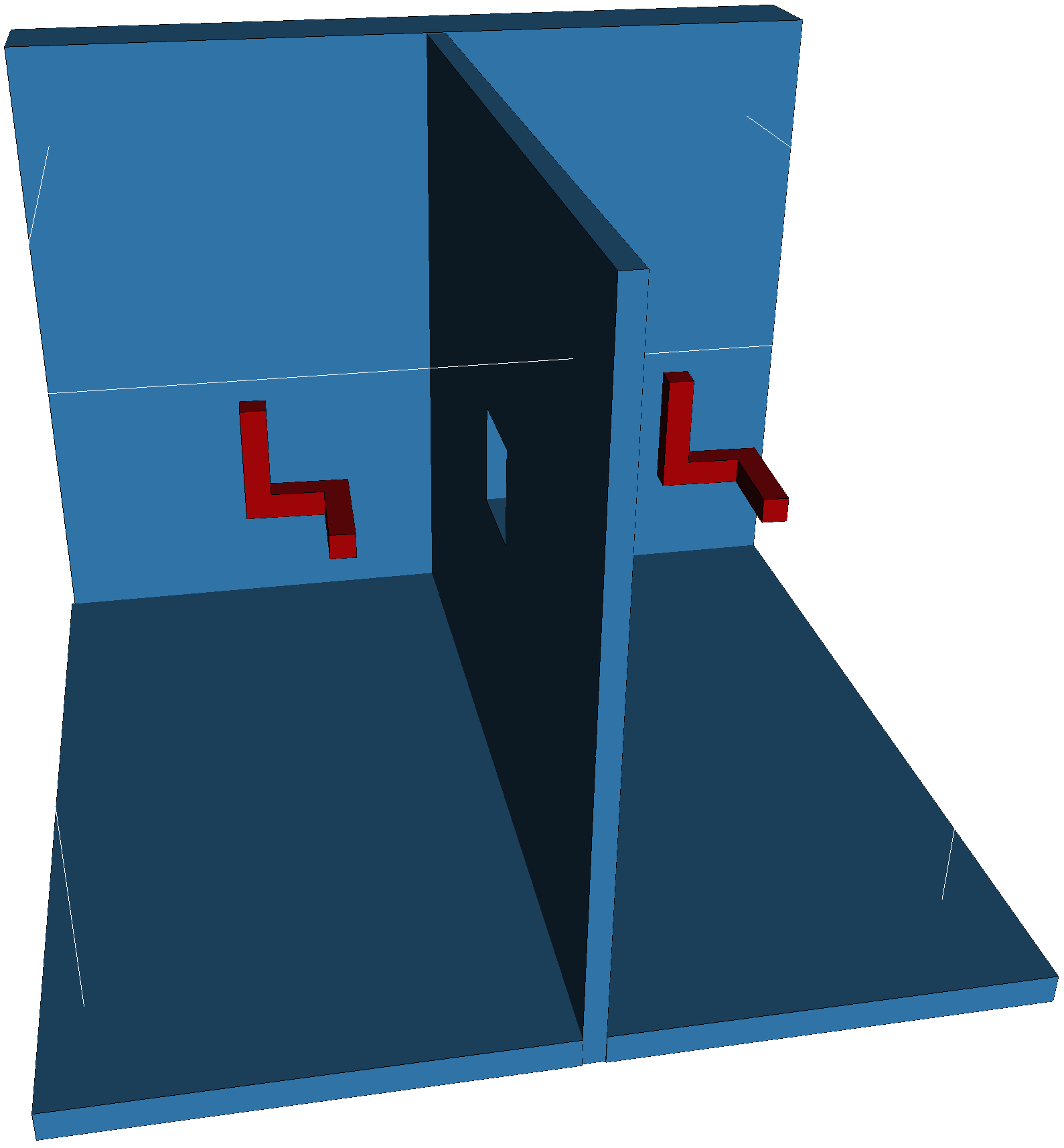}%
    \caption{\textbf{Motion Planning Scenarios}.  We run OMPL~\cite{OMPL} motion planning problems as benchmarks. Left-to-right: Apartment, Cubicles, Home, and Twistycool.  In these problems, the robot is a rigid-body object that must, through rotation and translation, find a collision-free path through the environment, from a start pose to a goal pose.}
    \label{fig:moption_planning_scenarios}
\end{figure}

\begin{table}[t]
    \centering
    \footnotesize
    \begin{tabular}{@{}lccc@{\quad}cc@{\quad}c@{}}\toprule
                  & Edge & Cloud & \multicolumn{2}{c}{\algname VPC} & \multicolumn{2}{c}{\algname Proxy} \\
                  \cmidrule(lr){4-5} \cmidrule(l){6-7}
         Scenario & Only &  Compute & Network & Total & Network & Total \\
         \midrule
         Apartment  & 157.6 &  {4.2} &  0.4 & \textbf{4.6} &   0.7 & \textbf{5.0} \\
         Cubicles   &  35.8 &  {1.4} &  0.3 & \textbf{1.7} &   0.6 & \textbf{2.1} \\
         Home       & 161.8 &  {6.2} &  0.3 & \textbf{6.5 }&   0.6 & \textbf{6.8} \\
         TwistyCool & 167.9 &  {5.1} &  0.4 & \textbf{5.5} &   0.6 & \textbf{5.7} \\
         \bottomrule
    \end{tabular}
    \caption{Multi-core Motion Planning with \algname.  \textup{We benchmark \algname on 6 different motion planning scenarios using the same multi-core motion planner, and record the compute time in seconds on the local edge computer, and the total(compute + network) time for using a 96-core computer in the cloud. \algname demonstrates up to 34.2x improvement with VPC and 31.52x improve with proxy than using only edge server. } }
    \label{tab:mp_times}
\end{table}

Motion planning computes a collision-free motion for a robot to get from one configuration to another.  Sampling-based motion planners randomly sample configurations and connect them together into a graph, rejecting samples and motions that are in collision. %
These planners can be %
scaled with additional computing cores.

Using \algname, we deploy a multi-core sampling-based motion planner~\cite{Ichnowski2014_TRO,ichnowski2019mpt} to a 96-core computer in the cloud to solve motion planning problems from the Open Motion Planning Library (OMPL)~\cite{OMPL} (see Fig.~\ref{fig:moption_planning_scenarios}). This planner node subscribes to topics for the collision model of the environment and motion plan requests
(Fig.~\ref{fig:mp_nodes}).  When the planner node receives a message on any of these topics, it computes a motion plan, and then publishes it to a separate topic. For more details on the multi-core motion planner, we refer the reader to the paper  and the open-source code by ~\citet{ichnowski2019mpt}.
To configure \algname to work with multi-core motion planner, we record the steps we use to setup the dependencies (e.g., FCL~\cite{pan2012fcl} and Nigh~\cite{ichnowski2018concurrent}) in a
script.
By providing the script, we configure \algname similar to Listing \ref{listing:launch}. %

We compare the planning time as the difference between publishing a motion plan request message, and receiving the plan result message, and show the results in Table~\ref{tab:mp_times}.  The same motion planning problem is solved in a fraction of the time on the cloud when compared to using the edge computer.  
If the motion planner is asymptotically-optimal %
(finds shorter/better plans the longer it runs and with more CPU cores), then one could potentially run the motion planner for the same amount of time but get a better path using the cloud.  \citet{anand2021serverless} explored and shown the benefit of using the tradeoff between more cores and the resulting motion plan optimality.

\section{Conclusion}

We present \algname, a user-friendly and adaptive extension to ROS that allows developers to rapidly deploy portions of their ROS system to computers in the cloud. \algname sets up a secure network channel transparent to the program code, allowing applications to be split between edge and the cloud with little to no modification. In experiments, we show that the added latency associated with pushing software components to the cloud is small when compared to the time gained from using high-end computers with many cores and GPUs in the cloud. However, in some simple tasks, using high-end cloud servers may lead to marginal benefits and can be considered as an overkill. 

In future work, we will address 
the interactions of multiple hardware systems with different ROS masters, and handle the decentralized communication efficiently and securely. 
We will also support real-time compression on the proxy connection between edge computer and cloud to help reduce latency especially on low-bandwidth connections. %

\section*{Acknowledgements}
This research was performed at the AUTOLAB at UC Berkeley in affiliation with the Berkeley AI Research (BAIR) Lab, and the CITRIS ``People and Robots" (CPAR) Initiative. %
The authors were supported in part by donations from Google, Siemens, Toyota Research Institute, Autodesk, Honda, Intel, Hewlett-Packard, and VMWare. This material is based upon work supported by the National Science Foundation Graduate Research Fellowship Program under Grant No. DGE 1752814 and NSF/VMware Partnership on Edge Computing
Data Infrastructure (ECDI), NSF award 1838833. Any opinions, findings, and conclusions or recommendations expressed in this material are those of the authors and do not necessarily reflect the views of the sponsors. We thank our colleagues who provided helpful feedback and suggestions, in particular Joseph M. Hellerstein and Alvin Cheung. %

\renewcommand*{\bibfont}{\footnotesize}
\printbibliography

\end{document}